# Deep Lambertian Networks

Yichuan Tang                                                      TANG@CS.TORONTO.EDU
Ruslan Salakhutdinov                                         RSALAKHU@CS.TORONTO.EDU
Geoffrey Hinton                                                HINTON@CS.TORONTO.EDU
Department of Computer Science, University of Toronto, Toronto, Ontario, CANADA

## Abstract

Visual perception is a challenging problem in part due to illumination variations. A possible solution is to first estimate an illumination invariant representation before using it for recognition. The object albedo and surface normals are examples of such representations. In this paper, we introduce a multilayer generative model where the latent variables include the albedo, surface normals, and the light source. Combining Deep Belief Nets with the Lambertian reflectance assumption, our model can learn good priors over the albedo from 2D images. Illumination variations can be explained by changing only the lighting latent variable in our model. By transferring learned knowledge from similar objects, albedo and surface normals estimation from a *single* image is possible in our model. Experiments demonstrate that our model is able to generalize as well as improve over standard baselines in *one-shot* face recognition.

## 1. Introduction

Multilayer generative models have recently achieved excellent recognition results on many challenging datasets (Ranzato & Hinton, 2010; Quoc et al., 2010; Mohamed et al., 2011). These models share the same underlying principle of first learning generatively from data before using the learned latent variables (features) for discriminative tasks. The advantage of using this indirect approach for discrimination is that it is possible to learn meaningful latent variables that achieve strong generalization. In vision, illumination is a major cause of variation. When the light source direction and intensity changes in a scene, dramatic changes in image intensity occur. This is detrimental to recognition performance as most algorithms use image intensities as inputs. A natural way of attacking this problem is to learn a model where the albedo, surface normals, and the lighting are explicitly represented as the latent variables. Since the albedo and surface normals are physical properties of an object, they are features which are invariant w.r.t. illumination.

Separating the surface normals and the albedo of objects using multiple images obtained under different lighting conditions is known as photometric stereo (Woodham, 1980). Hayakawa (1994) described a method for photometric stereo using SVD, which estimated the shape and albedo up to a linear transformation. Using integrability constraints, Yuille et al. (1999) proposed a similar method to reduce the ambiguities to a *generalized bas relief* ambiguity. A related problem is the estimation of intrinsic images (Barrow & Tenenbaum, 1978; Gehler et al., 2011). However, in those works, the shading (inner product of the lighting vector and the surface normal vector) instead of the surface normals is estimated. In addition, the use of three color channels simplifies that task.

In the domain of face recognition, Belhumeur & Kriegman (1996) showed that the set of images of an object under varying lighting conditions lie on a polyhedral cone (illumination cone), assuming a Lambertian reflectance and a fixed object pose. Recognition algorithms were developed based on the estimation of the illumination cone (Georghiades et al., 2001; Lee et al., 2005). The main drawback of these models is that they require multiple images of an object under varying lighting conditions for estimation. While Zhang & Samaras (2006); Wang et al. (2009) present algorithms that only use a single training image, their algorithms require bootstrapping with a 3D morphable face model. For every generic object class, building a 3D morphable model would be labor intensive.

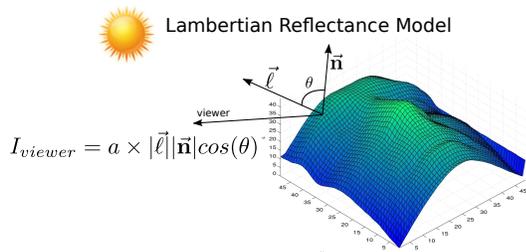

*Figure 1.* Diagram of the Lambertian Reflectance model. $\boldsymbol{\ell} \in \mathbb{R}^3$ points to the light source. $\vec{\mathbf{n}}_i \in \mathbb{R}^3$ is the surface normal, which is perpendicular to the tangent plane at a point on the surface.

In this paper, we introduce a generative model which **(a)** incorporates albedo, surface normals, and the lighting as latent variables; **(b)** uses multiplicative interaction to approximate the Lambertian reflectance model; **(c)** learns from sets of 2D images the distributions over the 3D object shapes; and **(d)** is capable of one-shot recognition from a single training example.

The Deep Lambertian Network (DLN) is a hybrid undirected-directed model with Gaussian Restricted Boltzmann Machines (and potentially Deep Belief Networks) modeling the prior over the albedo and surface normals. Good priors over the albedo and normals are necessary since for inference with a single image, the number of latent variables is 4 times the number of observed pixels. Estimation is an ill-posed problem and requires priors to find a unique solution. A density model of the albedo and the normals also allows for parameter sharing across individual objects that belong to the same class. The conditional distribution for image generation follows from the Lambertian reflectance model. Estimating the albedo and surface normals amounts to performing posterior inference in the DLN model with no requirements on the number of observed images. Inference is efficient as we can use alternating Gibbs sampling to approximately sample latent variables in the higher layers. The DLN is a *permutation invariant* model which can learn from any object class and strikes a balance between laborious approaches in vision (which require 3D scanning (Blanz & Vetter, 1999)) and the generic unsupervised deep learning approaches.

## 2. Gaussian Restricted Boltzmann Machines

We briefly describe the Gaussian Restricted Boltzmann Machines (GRBMs), which are used to model the albedo and surface normals. As the extension of binary RBMs to real-valued visible units, GRBMs (Hinton & Salakhutdinov, 2006) have been successfully applied to tasks including image classification, video action recognition, and speech recognition (Lee et al., 2009; Krizhevsky, 2009; Taylor et al., 2010; Mohamed et al., 2011). GRBMs can be viewed as a mixture of diagonal Gaussians with shared parameters, where the number of mixture components is exponential in the number of hidden nodes. With visible nodes $\mathbf{v} \in \mathbb{R}^{N_v}$ and hidden nodes $\mathbf{h} \in \{0,1\}^{N_h}$, the energy of the joint configuration is given by:

$$E_{GRBM}(\mathbf{v},\mathbf{h}) = \frac{1}{2}\sum_i \frac{(v_i-b_i)^2}{\sigma_i^2} - \sum_j c_j h_j - \sum_{ij} W_{ij} v_i h_j$$

The conditional distributions needed for inference and generation are given by:

$$p(h_j=1|\mathbf{v}) = \frac{1}{1+\exp(-\sum_i W_{ij}v_i - c_j)}, \quad (1)$$
$$p(v_i|\mathbf{h}) = \mathcal{N}(v_i|\mu_i,\sigma_i^2), \quad (2)$$

where $\mu_i = b_i + \sigma_i^2 \sum_j W_{ij} h_j$. Additional layers of binary RBMs are often stacked on top of a GRBM to form a Deep Belief Net (DBN) (Hinton et al., 2006). Inference in a DBN is approximate but efficient, where the probability of the higher layer states is a function of the lower layer states (see Eq. 1).

## 3. Deep Lambertian Networks

GRBMs and DBNs use Eq. 2 to generate the intensity of a particular pixel $v_i$. This generative model is inefficient when dealing with illumination variations in $\mathbf{v}$. Specifically, the hidden activations needed to generate a bright image of an object are very different from the activations needed to generate a dark image of the *same* object.

The Lambertian reflectance model is widely used for modeling illumination variations and is a good approximation for diffuse object surfaces (those without any specular highlights). Under the Lambertian model, illustrated in Fig. 1, the $i$-th pixel intensity is modelled as $v_i = a_i \times max(\vec{\mathbf{n}}_i^\top \vec{\boldsymbol{\ell}}, 0)$. The albedo $a_i$, also known as the reflection coefficient, is the diffuse reflectivity of a surface at pixel $i$, which is material dependent but illumination invariant. In contrast to the generative process of the GRBM, the image of an object under different lighting conditions can be generated without changing the albedo and the surface normals. Multiplications *within* hidden variables in the Lambertian model give rise to this nice property.

### 3.1. The Model

The DLN is a hybrid undirected-directed generative model that combines DBNs with the Lambertian re-flectance model. In the DLN, the visible layer consists of image pixel intensities $\mathbf{v} \in \mathbb{R}^{N_v}$, where $N_v$ is the number of pixels in the image. The first layer hidden variables are the albedo, surface normals, and a light source vector. Specifically, for every pixel $i$, there are two corresponding latent random variables: the albedo $a_i \in \mathbb{R}^1$ and surface normal $\mathbf{n}_i \in \mathbb{R}^3$. Over an image, $\mathbf{a} \in \mathbb{R}^{N_v}$ is the image albedo, $\mathbf{N}$ is the surface normals matrix of dimension $N_v \times 3$, where $\mathbf{n}_i$ denotes the $i$-th row of $\mathbf{N}$. The light source variable $\boldsymbol{\ell} \in \mathbb{R}^3$ points in the direction of the light source in the scene. We use GRBMs to model the albedo and surface normals, and a Gaussian prior to model $\boldsymbol{\ell}$. It is important to use GRBMs since we expect the distribution over albedo and surface normals to be multi-modal (see Fig. 4).

Fig. 2 shows the architecture of the DLN model: Panel (a) displays a standard network where filled triangles denote multiplicative gating between pixels and the first hidden layer. Panel (b) demonstrates the desired latent representations inferred by our model given input $\mathbf{v}$. While we use GRBMs as the prior models on albedo and surface normals, Deep Belief Network priors can be obtained by stacking additional binary RBM layers on top of the $g$ and $h$ layers. For clarity of presentation, in this section we use GRBM priors[1].

The DLN combines the elegant properties of the Lambertian model with the GRBMs, resulting in a deep model capable of learning albedo and surface normal statistics from images in a weakly-supervised fashion. The DLN has the following generative process:

$$p(\mathbf{v}, \mathbf{a}, \mathbf{N}, \boldsymbol{\ell}) = p(\mathbf{a})p(\mathbf{N})p(\boldsymbol{\ell})p(\mathbf{v}|\mathbf{a}, \mathbf{N}, \boldsymbol{\ell}) \quad (3)$$
$$p(\mathbf{a}) \sim GRBM(\mathbf{a})$$
$$p(\mathbf{N}) \approx GRBM(vec(\mathbf{N})) \quad (4)$$
$$p(\boldsymbol{\ell}) \sim \mathcal{N}(\boldsymbol{\ell}|\boldsymbol{\mu}_{\boldsymbol{\ell}}, \boldsymbol{\Lambda})$$
$$p(\mathbf{v}|\mathbf{a}, \mathbf{N}, \boldsymbol{\ell}) = \prod_i^{N_v} \mathcal{N}(v_i|a_i(\mathbf{n}_i^\mathsf{T}\boldsymbol{\ell}); \sigma_{v_i}^2), \quad (5)$$

where $vec(\mathbf{N})$ denotes the vectorization of matrix $\mathbf{N}$.

The GRBM prior in Eq. 4 is only approximate since we enforce the soft constraint that the norm of $\mathbf{n}_i$ is equal to 1.0. We achieve this via an extra energy term in Eq. 6. Eq. 5 represents the probabilistic version of the Lambertian reflectance model. We have dropped "$max$" for convenience. "$max$" is not critical in our model as maximum likelihood learning regulates the generation process. In addition, a prior on lighting direction fits well with the psychophysical observations that human perception of shape relies on the assump-tion that light originates from above (Kleffner & Ramachandran, 1992).

---
[1]Extending our model to more flexible DBN priors is trivial.

DLNs can also handle multiple images of the same object under varying lighting conditions. Let $P$ be the number of images of the same object. We use $\mathbf{L} \in \mathbb{R}^{3 \times P}$ to represent the lighting matrix with columns $\{\boldsymbol{\ell}_p : p = 1, 2, \ldots, P\}$, and $\mathbf{V} \in \mathbb{R}^{N_v \times P}$ to represent the matrix of corresponding images. The DLN energy function is defined as:

$$E_{DLN}(\mathbf{V}, \mathbf{a}, \mathbf{N}, \mathbf{L}, \mathbf{g}, \mathbf{h}) = \frac{1}{2}\sum_p^P \sum_i^{N_v} \frac{(v_{ip} - a_i(\mathbf{n}_i^\mathsf{T}\boldsymbol{\ell}_p))^2}{\sigma_{v_i}^2}$$
$$+ \frac{1}{2}\sum_p^P (\boldsymbol{\ell}_p - \boldsymbol{\mu}_l)^\mathsf{T}\boldsymbol{\Lambda}(\boldsymbol{\ell}_p - \boldsymbol{\mu}_l) + \frac{\eta}{2}\sum_i^{N_v}(\mathbf{n}_i^\mathsf{T}\mathbf{n}_i - 1.0)^2$$
$$+ E_{GRBM}(\mathbf{a}, \mathbf{h}) + E_{GRBM}(vec(\mathbf{N}), \mathbf{g}) \quad (6)$$

The first line in the energy function is proportional to $\log p(\mathbf{v}|\mathbf{a}, \mathbf{N}, \boldsymbol{\ell})$, the multiplicative interaction term from the Lambertian model. The second line corresponds to the quadratic energy of $\log p(\boldsymbol{\ell})$ and the soft norm constraint on $\mathbf{n}_i$. This constraint is critical for the correct estimation of the albedo, since we can interpret the albedo at each pixel as the $L_2$ norm of the pixel surface normal. The third line contains the two GRBM energies: $\mathbf{h} \in \mathbb{R}^{N_h}$ represents the binary hidden variables of the albedo GRBM and $\mathbf{g} \in \mathbb{R}^{N_g}$ represents the hiddens of the surface normal GRBM:

$$E_{GRBM}(\mathbf{a}, \mathbf{h}) =$$
$$\frac{1}{2}\sum_i^{N_v} \frac{(a_i - b_i)^2}{\sigma_{a_i}^2} - \sum_j^{N_h} c_j h_j - \sum_{i,j}^{N_v, N_h} W_{ij} a_i h_j \quad (7)$$

$$E_{GRBM}(vec(\mathbf{N}), \mathbf{g}) = \frac{1}{2}\sum_{i,m=1}^{N_v, 3} \frac{n_{im}^2}{\sigma_{n_{im}}^2} - \sum_{i,m=1}^{N_v, 3} \frac{d_{im} n_{im}}{\sigma_{n_{im}}^2}$$
$$- \sum_k^{N_g} e_k g_k - \sum_{i,m=1,k}^{N_v, 3, N_g} U_{imk} n_{im} g_k \quad (8)$$

### 3.2. Inference

Given images of the same object under one or more lighting conditions, we want to infer the posterior distribution over the latent variables (including albedo, surface normals and light source): $p(\mathbf{a}, \mathbf{N}, \mathbf{L}, \mathbf{g}, \mathbf{h}|\mathbf{V})$. With GRBMs modeling the albedo $\mathbf{a}$ and surface normals $\mathbf{N}$, the posterior is complicated with no closed form solution. However, we can resort to Gibbs sampling using 4 sets of conditional distributions:

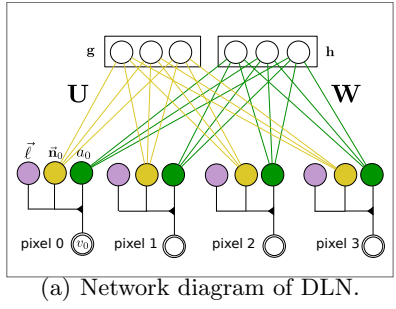

(a) Network diagram of DLN.

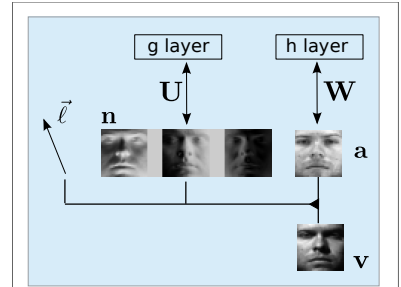

(b) Face images to illustrate the DLN.

*Figure 2.* Graphical model of the Deep Lambertian Network. The yellow weights model the surface normals while the green weights model the albedo. The arrow in the left figure is the light source direction vector, pointing towards the light source. Note that the light vector is shared for all pixels in the image. Best viewed in color.

- *Conditional 1*: $p(\mathbf{g}, \mathbf{h}|\mathbf{a}, \mathbf{N}, \mathbf{L}, \mathbf{V})$
- *Conditional 2*: $p(\mathbf{a}|\mathbf{N}, \mathbf{L}, \mathbf{h}, \mathbf{V})$
- *Conditional 3*: $p(\mathbf{L}|\mathbf{N}, \mathbf{a}, \mathbf{V})$
- *Conditional 4*: $p(\mathbf{N}|\mathbf{a}, \mathbf{L}, \mathbf{g}, \mathbf{V})$

*Conditional 1* is easy to compute as it factorizes over $\mathbf{g}$, and $\mathbf{h}$: $p(\mathbf{g}, \mathbf{h}|\mathbf{a}, \mathbf{N}, \mathbf{L}, \mathbf{v}) = p(\mathbf{h}|\mathbf{a})p(\mathbf{g}|\mathbf{N})$. Since Gaussian RBMs model the albedo $\mathbf{a}$ and the surface normals $\mathbf{N}$, the two factorized conditional distributions have the same form as Eq. 1.

*Conditional 2* factorizes into a product of Gaussian distributions over $N_v$ pixel-specific albedo variables:

$$p(\mathbf{a}|\mathbf{N}, \mathbf{L}, \mathbf{h}, \mathbf{V}) = \prod_i^{N_v} p(a_i|\mathbf{N}, \mathbf{L}, \mathbf{h}, \mathbf{V}) \sim$$

$$\prod_i^{N_v} \mathcal{N}\Big(a_i \Big| \frac{\sigma_{a_i}^2 \sum_p s_{ip} v_{ip} + \phi_i^h \sigma_{v_i}^2}{\sigma_{a_i}^2 \sum_p s_{ip}^2 + \sigma_{v_i}^2}; \frac{\sigma_{a_i}^2 \sigma_{v_i}^2}{\sigma_{a_i}^2 \sum_p s_{ip}^2 + \sigma_{v_i}^2}\Big),$$

where $s_{ip} = \mathbf{n}_i^\mathsf{T} \boldsymbol{\ell}_p$ is the illumination shading at pixel $i$ and $\phi_i^h = b_i + \sigma_{a_i}^2 \sum_j W_{ij} h_j$ is the top-down influence of the albedo GRBM.

This conditional distribution has a very intuitive interpretation. When a light source has zero strength, ($\boldsymbol{\ell}_p = 0 \to s_{ip} = 0$), then $p(a_i|\mathbf{n}_i, \boldsymbol{\ell}_p, \mathbf{h}, v_i)$ has mean at $\phi_i^h$, which is purely the top-down activation.

*Conditional 3* factorizes into a product distribution over $P$ separate light variables: $p(\mathbf{L}|\mathbf{N}, \mathbf{a}, \mathbf{V}) = \prod_{p=1}^P p(\boldsymbol{\ell}_p|\mathbf{N}, \mathbf{a}, \mathbf{v}_p)$, where $p(\boldsymbol{\ell}_p|\mathbf{N}, \mathbf{a}, \mathbf{v}_p)$ is defined by a quadratic energy function:

$$E(\boldsymbol{\ell}_p|\mathbf{N}, \mathbf{a}, \mathbf{v}) = \frac{1}{2}\boldsymbol{\ell}_p^\mathsf{T}\Big(\boldsymbol{\Lambda} + \sum_i \frac{\mathbf{m}_i \mathbf{m}_i^\mathsf{T}}{\sigma_{v_i}^2}\Big)\boldsymbol{\ell}_p$$
$$- \Big(\boldsymbol{\mu}_l^\mathsf{T}\boldsymbol{\Lambda} + \sum_i \frac{v_{ip}\mathbf{m}_i}{\sigma_{v_i}^2}\Big)^\mathsf{T}\boldsymbol{\ell}_p.$$

Hence the conditional distribution over $\boldsymbol{\ell}_p$ is a multivariate Gaussian of the form:

$$p(\boldsymbol{\ell}_p|\mathbf{N}, \mathbf{a}, \mathbf{v}) \sim \mathcal{N}(\boldsymbol{\ell}_p|\tilde{\boldsymbol{\Lambda}}^{-1}\Big(\boldsymbol{\mu}_l^\mathsf{T}\boldsymbol{\Lambda} + \sum_i \frac{v_{ip}\mathbf{m}_i}{\sigma_{v_i}^2}\Big); \tilde{\boldsymbol{\Lambda}}^{-1}),$$

where $\tilde{\boldsymbol{\Lambda}} = \boldsymbol{\Lambda} + \sum_i \frac{\mathbf{m}_i \mathbf{m}_i^\mathsf{T}}{\sigma_{v_i}^2}$, and $\mathbf{m}_i = a_i \mathbf{n}_i$.

*Conditional 4* can be decomposed into a product of distributions over the surface normals of each pixel:

$$p(\mathbf{N}|\mathbf{L}, \mathbf{a}, \mathbf{g}, \mathbf{V}) = \prod_i p(\mathbf{n}_i|\mathbf{L}, \mathbf{g}, a_i, \mathbf{v}_i)$$

Since in our model we have the soft norm constraint on $\mathbf{n}_i$ ( $\frac{\eta}{2}\sum_i^{N_v}(\mathbf{n}_i^\mathsf{T}\mathbf{n}_i - 1.0)^2$ ), there is no simple closed form for $p(\mathbf{n}_i|\mathbf{L}, \mathbf{g}, a_i, \mathbf{v}_i)$. We use the Hamiltonian Monte Carlo (HMC) algorithm for sampling.

HMC (Duane et al., 1987; Neal, 2010) is an auxiliary variable MCMC method which combines Hamiltonian dynamics with the Metropolis algorithm to sample continuous random variables. In order to use HMC, we must have a differentiable energy function over the variables. In this case, the energy of conditional 4 takes form:

$$E(\mathbf{n}_i) = \frac{1}{2}\mathbf{n}_i^\mathsf{T}\left(\frac{\sum_p (a_i \boldsymbol{\ell}_p)(a_i \boldsymbol{\ell}_p)^\mathsf{T}}{\sigma_{v_i}^2} + \mathbf{D}_i\right)\mathbf{n}_i$$
$$- \left(\frac{a_i \sum_p v_{ip}\boldsymbol{\ell}_p}{\sigma_{v_i}^2} + \boldsymbol{\phi}_i^{g\mathsf{T}}\mathbf{D}_i\right) + \frac{\eta}{2}(\mathbf{n}_i^\mathsf{T}\mathbf{n}_i - 1)^2,$$

where $\boldsymbol{\phi}_i^g$ is the top-down mean of $\mathbf{n}_i$ from the g-layer, and $\mathbf{D}_i = diag(\sigma_{n_{i1}}^{-2}, \sigma_{n_{i2}}^{-2}, \sigma_{n_{i3}}^{-2})$ is the $3 \times 3$ diagonal matrix.

We note that there is a linear ambiguity when we estimate the normals and lighting direction. In Eq. 5, $\mathbf{n}_i^\mathsf{T}\boldsymbol{\ell}_p = \mathbf{n}_i^\mathsf{T}\mathbf{R}\mathbf{R}^{-1}\boldsymbol{\ell}_p$. This means that we can only estimate $\mathbf{n}_i$ and $\boldsymbol{\ell}_p$ up to a linear transformation. Fortunately, while $\mathbf{R}$ is unknown, it is *constant* across $\{\mathbf{v}_i\}_{i=1}^{P}$ due to the learned priors over $\mathbf{N}$, $\mathbf{a}$ and $\boldsymbol{\ell}$. Therefore, recognition and image relighting tasks (Sec. 4) are not affected.

### 3.3. Learning

Learning is accomplished using a variant of the EM algorithm. In the E-step, MCMC samples are drawn from the approximate posterior distribution (Neal & Hinton, 1998). We first sample from the conditional distributions in Sec. 3.2 to approximate the posterior $p(\mathbf{a},\mathbf{N},\mathbf{L},\mathbf{h},\mathbf{g}|\mathbf{V};\theta^{old})$. We then optimize the joint log-likelihood function w.r.t. the model parameters. Specifically,

$$\triangle\theta = -\alpha\int p(\mathbf{a},\mathbf{N},\mathbf{L},\mathbf{h},\mathbf{g}|\mathbf{V};\theta^{old})\frac{\partial}{\partial\theta}\big\{E(\mathbf{V},\mathbf{a},\mathbf{N},\mathbf{L},\mathbf{h},\mathbf{g};\theta)\big\} \quad (9)$$

where $\alpha$ is the learning rate. We approximate the integral using:

$$\frac{1}{N}\sum_{i}^{N}\frac{\partial}{\partial\theta}\big\{E(\mathbf{V},\mathbf{a}^{(i)},\mathbf{N}^{(i)},\mathbf{L}^{(i)},\mathbf{h}^{(i)},\mathbf{g}^{(i)};\theta)\big\},$$

where the samples $\{\mathbf{a}^{(i)},\mathbf{N}^{(i)},\mathbf{L}^{(i)},\mathbf{h}^{(i)},\mathbf{g}^{(i)}\}$ are approximately drawn from the posterior distribution $p(\mathbf{a},\mathbf{N},\mathbf{L},\mathbf{h},\mathbf{g}|\mathbf{V};\theta^{old})$ in the E-step. Maximum likelihood learning of GRBMs (and DBNs) is intractable. We therefore turn to Contrastive Divergence (CD) (Hinton, 2002) to compute an approximate gradient during learning. The complete training algorithm for the DLN in presented in Alg. 1.

Rather than starting with randomly initialized weights, we can achieve better convergence by first training the albedo GRBM on a separate face database. We can then transfer the learned weights before learning the complete DLN.

## 4. Experiments

We experiment with the Yale B and the Extended Yale B face databases. Combined, the two databases contain 64 frontal images of 38 different subjects. 45 images for each subject are further divided into 4 subsets of increasing illumination variations. Fig. 3 shows samples from the Yale B and Extended Yale B database.

For each subject, we used approximately 45 frontal images for our experiments[2]. We separated 28 subjects from the Extended Yale B database for training and held-out all 10 subjects from the original Yale B database for testing[3]. The preprocessing step involved

---
[2]A few of the images are corrupted.
[3]We used the cropped images provided by the Yale B

**Algorithm 1** Learning Deep Lambertian Networks
1: Pretrain the $\{\mathbf{a},\mathbf{h}\}$ albedo GRBM with faces images and initialize $\{\mathbf{W},\mathbf{b},\mathbf{c}\}$ of the DLN with the GRBM's parameters.
2: Initialize other weights $\sim\mathcal{N}(0,0.01^2)$, $\sigma^2\leftarrow 1.0$.
**repeat**
 //*Approximate E-step*:
 **for** $n=1$ **to** #training subjects **do**
3:  Given $\mathbf{V}^n$, sample $p(\mathbf{a},\mathbf{N},\mathbf{L},\mathbf{h},\mathbf{g}|\mathbf{V^n};\theta^{old})$ using the conditionals defined in Sec. 3.2, obtaining samples of $\{\mathbf{a}^{(i)},\mathbf{N}^{(i)},\mathbf{L}^{(i)}\}$.
 **end for**

 //*Approximate M-step*:
4: Treating $\{\mathbf{a}^{(i)}\}$ as training data, CD is used to learn the weights of the albedo GRBM.
5: Treating $\{\mathbf{N}^{(i)}\}$ as training data, CD is used to learn the weights of the surface normal GRBM.
6: Maximum likelihood estimations of the parameters $\sigma^2_{v_i}$, $\boldsymbol{\mu}_{\boldsymbol{\ell}}$, and $\boldsymbol{\Lambda}$ are computed.
**until** convergence

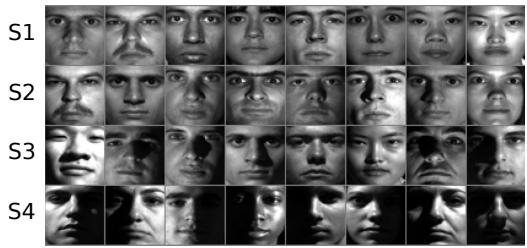

*Figure 3.* Examples from the Yale B Extended face database. Each row contains samples from an illumination subset.

downsizing the face images to the resolution of $24\times 24$. Using equations of Sec. 3.2, we can infer one albedo image and one set of surface normals from each of the 28 subjects. These 28 training albedo and surface normal samples are insufficient for multilayer generative models with millions of parameters. Therefore, we leverage a large set of the face images from the Toronto Face Database (TFD) (Susskind et al., 2011). The TFD is a collection of 100,000 face images from a variety of other datasets. To create more training data for the surface normals, we randomly translated all 28 sets of them by $\pm 2$ pixels.

The DLN used 2 layer DBNs (instead of single layer GRBMs) to model the priors over $\mathbf{a}$ and $\mathbf{N}$. The albedo DBN had 800 $\mathbf{h}^1$ nodes and 200 $\mathbf{h}^2$ nodes. The normals DBN had 1000 $\mathbf{g}^1$ nodes and 100 $\mathbf{g}^2$ nodes. To see what the DLN's prior on the albedo looks like, we show samples generated by the albedo DBN in Fig. 4.

---
Extended database website: http://goo.gl/LKwtX.

Deep Lambertian Networks

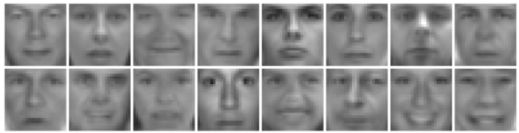

Figure 4. Random samples after 50,000 Gibbs iterations of the Deep Belief Network modeling the learned albedo prior.

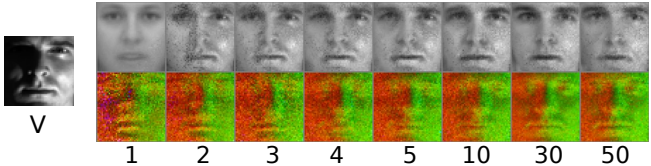

Figure 5. **Left**: A single input test image. **Right**: Intermediate samples during alternating Gibbs sampling: iterations 1 to 50. Top row contains the estimated albedo. Bottom row contains the estimated surface normals. The albedo and surface normal were initialized with the visible biases of their respective GRBMs. Best viewed in color.

Learning the multi-modal albedo prior is made possible by the use of unsupervised TFD data.

### 4.1. Inference

After learning, we investigated the inference process in the DLN. Although the DLN can use multiple images of the same object during inference, it is important to investigate how well it performs with a *single* test image. We are also interested in the number of iterations that sampling would take to find the posterior modes.

In our first experiment, we presented the model with a single Yale B face image from a *held-out* test subject, as shown in Fig. 5. The light source illuminates the subject from the bottom right, causing a significant shadow across the top left of the subject's face. Since the albedo captures a lighting invariant representation of the face, correct posterior distribution should automatically perform illumination normalization. Using the algorithm described in Sec. 3.2, we clamp the visible nodes to the test face image and sample from the 4 conditionals in an alternating fashion. HMC was used to sample from $\mathbf{N}$. In total, we perform 50 iterations of alternating Gibbs sampling. During each iteration, the $\mathbf{N}$ variables are sampled using HMC with 20 leapfrog iterations and 10 HMC epochs. The step size was set to 0.01 with a momentum of 2.0. The acceptance rate was around 0.7.

We plot the intermediate samples from iterations 1 to 50 in Fig. 5. The top row displays the inferred albedo $\mathbf{a}$. At every pixel, there is a surface normal vector $\mathbf{n}_i \in \mathbb{R}^3$. For visual presentation, we treat each $\mathbf{n}_i$ as a RGB pixel and plot them as color images in the bottom row. Note that the Gibbs chain quickly jumps (at iteration 5) into the correct mode. Good results are obtained due to the knowledge transfer of the albedo and surface normals learned from other subjects.

We next randomly selected single test images from the 10 Yale B test subjects. Using exactly the same sampling algorithm, Fig. 6(a) shows their inferred albedo and surface normals. The first column displays the test image, the middle and right columns contain the estimated albedo and surface normals. We also found that using two test images per subject improves performance. Specifically, we sampled from $p(\mathbf{a}, \mathbf{N}|\mathbf{V} \in \mathbb{R}^{N_v \times 2})$ instead of $p(\mathbf{a}, \mathbf{N}|\mathbf{v} \in \mathbb{R}^{N_v})$. The results are displayed in Fig. 6(b).

### 4.2. Relighting

The task of face relighting is useful to demonstrate *strong generalization* capabilities of the model. The goal is to generate face images of a particular person under never-before seen lighting conditions. Realistic images can only be generated if the albedo and surface normals of that particular person were correctly inferred. We first sample the lighting variable $\boldsymbol{\ell}$ from its Gaussian prior defined by $\{\boldsymbol{\mu}, \boldsymbol{\Lambda}\}$. Conditioned on the inferred $\mathbf{a}$ and $\mathbf{N}$ (see Fig. 6(b)), we use Eq. 5 to draw samples of $\mathbf{v}$. Fig. 6(c) shows relighted face images of *held-out* test subjects.

### 4.3. Recognition

We next test the performance of DLN at the task of face recognition. For the 10 test subjects of Yale B, only image(s) from subset 1 (with 7 images) are used for training. Images from subsets 2-4 are used for testing. In order to use DLN for recognition, we first infer the albedo ($a_i$) and surface normals ($\mathbf{n}_i$) conditioned on the provided training image(s) of test subjects. For every subject, a 3 dimensional linear subspace is spanned by the inferred albedo and surface normals. In particular, we consider the matrix $\mathbf{M}$ of dimensions $N_v \times 3$, with the $i$-th row set to $\mathbf{m}_i = a_i \mathbf{n}_i$. The columns of $\mathbf{M}$ spans the 3 dimensional linear subspace. Test images of the test subjects are compared to all 10 subspaces and are labeled according to the label of its nearest subspace.

Fig. 7 plots the recognition errors as a function of number of training images used. DBN results are obtained by training a 2 layer DBN directly on the training images, and a linear SVM was trained on the top-most hidden activations of the DBN. That standard DBN can not handle lighting variations very accurately. Another approach, called Normalized Correlation, first normalizes images to unit norm. For each test image, its cosine similarity to all training images is computed. The test image takes on the label of the closest training image. Normalized Correlation performs significantly better than Nearest Neighbor due to its normalization, which removes some of the lighting variations. Finally, the SVD method finds a 3 dimensional linear subspace (with the largest singular values) spanned by the training images of each of the test subjects. A test image is assigned to the closest subspace.

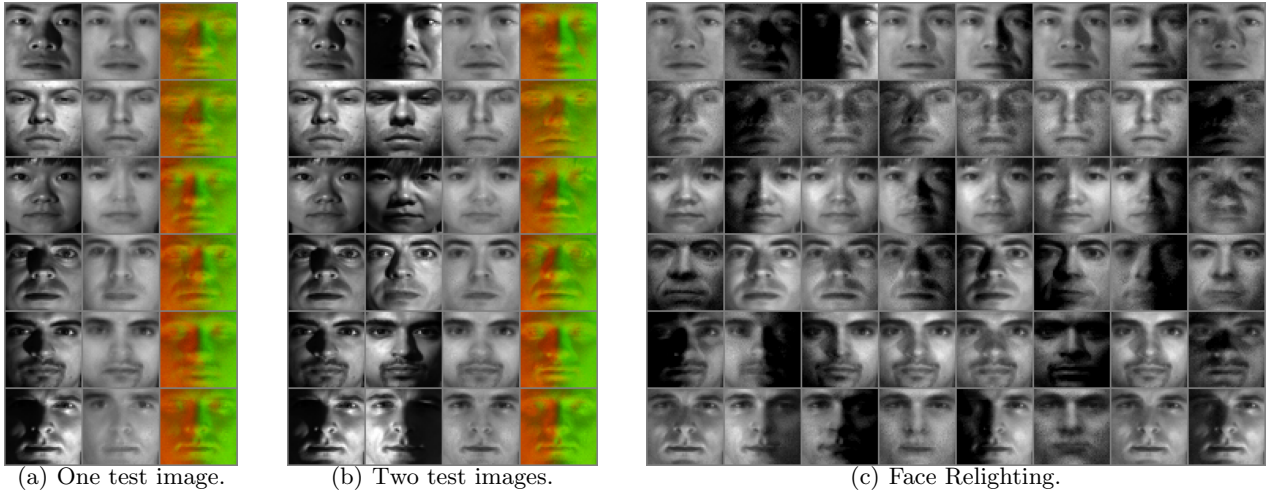

(a) One test image.  (b) Two test images.  (c) Face Relighting.

*Figure 6.* **Left**: Inference results when using only a single test image. 1st column is the test images, 2nd column is the albedo and the 3rd column is the surface normals. **Middle**: Results improve slightly when using an additional test image with a different illumination. **Right**: Using the estimated albedo and surface normals, we show synthesized images under novel lighting conditions. Best viewed in color.

We note that for the important task of *one-shot* recognition, DLN significantly outperforms many other methods. In the computer vision literature, Zhang & Samaras (2006); Wang et al. (2009) report lower error rates on the Yale B dataset. However, their algorithms make use of pre-existing 3D morphable models, whereas the DLN learns the 3D information automatically from 2D images.

### 4.4. Generic Objects

The DLN is applicable not only on face images but also images of generic objects. We used 50 objects from the Amsterdam Library of Images (ALOI) database (Geusebroek et al., 2005). For every object, 15 images of varying lighting were divided into 10 for training and 5 for testing. Using the provided masks for each object, images are cropped and rescaled to the resolution of $48 \times 48$. We used a DLN with $N_h = 1000$ and $N_g = 1500$. A 500 $\mathbf{h}^2$ layer and 500 $\mathbf{g}^2$ layer were also added. After training, we performed posterior inference using one of the held-out image. Fig. 8 shows results. The top row contains test images, the middle row displays the inferred albedo images after 50 alternating Gibbs iterations, and the bottom row shows the inferred surface normals.

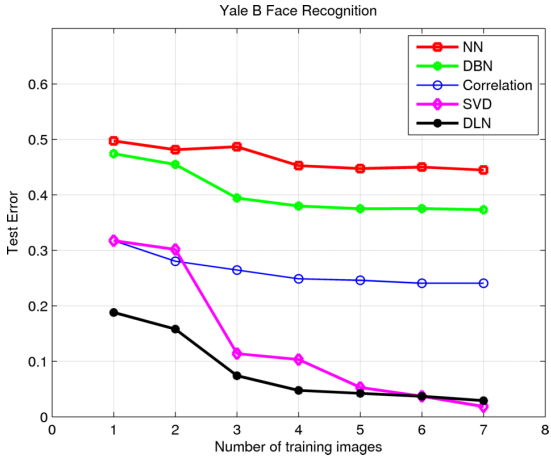

*Figure 7.* Recognition results on the Yale B face database. **NN**: nearest neighbor. **DBN**: Deep Belief Network. **Correlation**: normalized cross correlation. **SVD**: singular value decomposition. **DLN**: Deep Lambertian Network.

## 5. Discussions

We have introduced a generative model with meaningful latent variables and multiplicative interactions simulating the Lambertian Reflectance model. We have shown that by learning priors on these illumination-invariant variables directly from data, we can improve on one-shot recognition tasks as well as generate images under novel illuminations.

# Deep Lambertian Networks

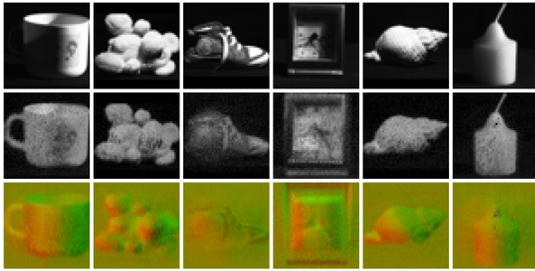

*Figure 8.* Inference conditioned on test objects, using 50 Gibbs iterations. **Top**: Images of objects under new illumination. **Middle**: Inferred albedo. **Bottom**: Inferred surface normals.

## Acknowledgements
We thank Maksims Volkovs, James Martens, and Abdelrahman Mohamed for discussions. This research was supported by NSERC and CIFAR.